\documentclass[runningheads]{llncs}

\usepackage[english]{babel}
\usepackage{amsmath,amssymb}
\usepackage{graphicx}
\usepackage[colorlinks=true, allcolors=blue]{hyperref}
\usepackage{url} 

\usepackage{booktabs}      
\usepackage{subcaption}    
\usepackage{multirow}
\usepackage[export]{adjustbox}
\usepackage{enumitem}
\usepackage{pifont}
\usepackage{array}
\usepackage{booktabs}  
\usepackage[table]{xcolor} 
\usepackage{graphicx}  
\usepackage{array}     
\usepackage{adjustbox}
\usepackage{makecell}
\usepackage[colorlinks=true, allcolors=blue]{hyperref}
\hypersetup{pdftitle={}, pdfauthor={}, pdfsubject={}, pdfkeywords={}}
\usepackage{url}
\usepackage{hyperref}  
\hypersetup{
    colorlinks=true,
    linkcolor=blue,
    filecolor=magenta,      
    urlcolor=blue, 
    citecolor=blue,
}
\usepackage{bbding}
\definecolor{lightcyan}{rgb}{0.88, 0.95, 1.0}

\title{LLM-MRD: LLM-Guided Multi-View Reasoning Distillation for Fake News Detection}

\author{Weilin Zhou\inst{1,3} \and
Shanwen Tan\inst{2} \and
Enhao Gu\inst{1,3} \and
Yurong Qian\inst{1,3}\textsuperscript{(\Envelope)}} 

\authorrunning{W. Zhou et al.}

\institute{Xinjiang University, Urumqi, China \and
Sichuan University, Chengdu, China \and
Joint International Research Laboratory of Silk Road Multilingual Cognitive Computing, Xinjiang University, Urumqi, China\\
\email{zhouweilin55@gmail.com, 2023141010009@stu.scu.edu.cn, Guenhao2005@163.com, qyr@xju.edu.cn}}

\titlerunning{LLM-MRD}

\begin{document}
\maketitle

\begin{abstract}
Multimodal fake news detection is crucial for mitigating societal disinformation. Existing approaches attempt to address this by fusing multimodal features or leveraging Large Language Models (LLMs) for advanced reasoning. However, these methods suffer from serious limitations, including a lack of comprehensive multi-view judgment and fusion, and prohibitive reasoning inefficiency due to the high computational costs of LLMs. To address these issues, we propose \textbf{LLM}-Guided \textbf{M}ulti-View \textbf{R}easoning \textbf{D}istillation for Fake News Detection ( \textbf{LLM-MRD}), a novel teacher-student framework. The Student Multi-view Reasoning module first constructs a comprehensive foundation from textual, visual, and cross-modal perspectives. Then, the Teacher Multi-view Reasoning module generates deep reasoning chains as rich supervision signals. Our core Calibration Distillation mechanism efficiently distills this complex reasoning-derived knowledge into the efficient student model. Experiments show LLM-MRD significantly outperforms state-of-the-art baselines. Notably, it demonstrates a comprehensive average improvement of 5.19\% in ACC and 6.33\% in F1-Fake when evaluated across all competing methods and datasets. Our code is available at https://anonymous.4open.science/r/LLM-MRD-CC50
\end{abstract}


\section{Introduction} The proliferation of social media has accelerated the dissemination of fake news, often manipulated for political or economic gain\cite{liu2024fakenewsgpt4,qiu2025dsen}. This phenomenon threatens social stability, rendering automatic detection a critical research imperative. However, existing detection methods face significant challenges. Models trained on specific domains (e.g., politics, health) often fail to generalize to real-world, multi-domain news streams\cite{wei2022chain}. Furthermore, the prevalence of multimodal content (text-image) introduces further complexity\cite{comito2023multimodal}, as many approaches struggle to perform holistic analysis or achieve accurate semantic interpretation.

Current fake news detection methods can be broadly categorized into three groups. The first, unimodal methods, analyze a single modality, such as textual cues in DSTnet\cite{chakraborty2022detecting} or visual artifacts in {MVAN\cite{ni2021mvan}. Such methods, however, inherently fail to capture cross-modal inconsistencies. The second, multimodal fusion methods, combine information from multiple modalities, evolving from simple concatenation in SpotFake\cite{singhal2019spotfake} to co-attention in HMCAN and consistency modeling in CAFE\cite{chen2022cross}. Despite this, they often yield incomplete or noisy fused representations. The third category leverages Large Language Models (LLMs) for advanced reasoning, as in LLM-GAN\cite{wang2024llm} and DIFND\cite{yan2025debunk}, yet they typically rely on single-point judgments or naive aggregation, lacking robust collaborative fusion mechanisms.

Despite these innovations, methods in all three categories suffer from fundamental weaknesses: poor cross-modal understanding, insufficient fusion strategies, and ineffective evidence integration, even when assisted by LLMs. These deficiencies inhibit models from fully capturing the news context, rendering them vulnerable to sophisticated misinformation. The primary limitations are:

\begin{enumerate}[label=\arabic*., leftmargin=*, start=1]
\item \textbf{Insufficient Multi-View Judgment and Fusion:} Existing methods often fail to analyze content from comprehensive and distinct perspectives (e.g., textual, visual, and cross-modal). Furthermore, even when different views are considered, they often lack robust mechanisms to effectively fuse the reasoning and evidence derived from these different angles\cite{singhal2019spotfake,ren2024mmsfd}, leading to an incomplete or contradictory understanding.

\item \textbf{Prohibitive Reasoning Inefficiency:} While large language models (LLMs) offer advanced reasoning capabilities, their direct application in detection systems is often impractical due to high computational costs and significant inference latency. To overcome this, knowledge distillation is employed to transfer knowledge from a large teacher model to a efficient student model, enabling efficient deployment without sacrificing reasoning quality\cite{wang2024llm,liu2024stepwise}.
\end{enumerate}

\indent To address these limitations, we derive the \textbf{LLM-Guided Multi-View Reasoning Distillation for Fake News Detection} (LLM-MRD) framework. This novel teacher-student approach is designed to leverage a powerful LLM as an expert teacher performing multi-perspective reasoning over text, images, and their interactions. This complex reasoning is then transferred to a efficient student model via calibrated distillation. This introduces an expert-guided reasoning distillation paradigm, shifting the focus from feature fusion to hierarchical reasoning-knowledge transfer. An LLM expert teacher generates deep, multi-faceted reasoning, which is subsequently distilled into the student model using a specialized calibration mechanism, enhancing its robustness and semantic understanding. Our main contributions are:

\begin{enumerate}[label=\arabic*., leftmargin=*, start=1] \item We propose a Multi-Perspective Student Architecture designed to mitigate information silos in unimodal analysis. We constructed this module to establish a comprehensive reasoning foundation by extracting complementary features from textual (BERT\cite{devlin2019bert}), visual (MAE\cite{he2022masked}), and cross-modal (CLIP\cite{radford2021learning}) views.

\begin{figure}[htbp]
    \centering
    \begin{subfigure}{\columnwidth}
        \centering
        \includegraphics[width=0.7\columnwidth]{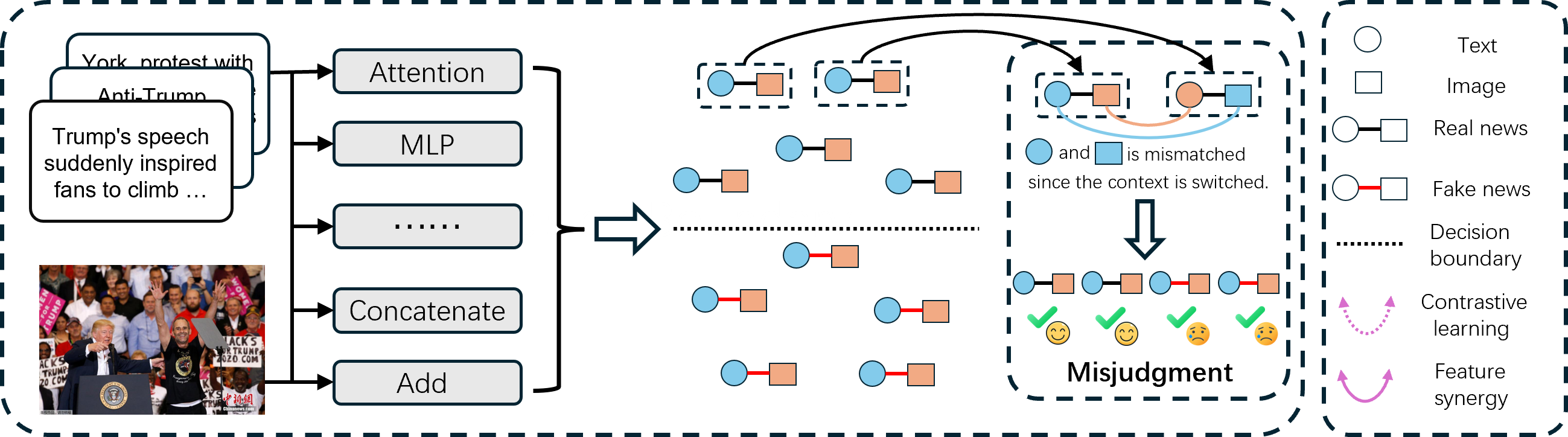}
        \caption{Previous methods}
        \label{fig:sub_distillation_and_temp}
    \end{subfigure}
    \vspace{0.15cm} 
    \begin{subfigure}{\columnwidth}
        \centering
        \includegraphics[width=0.7\columnwidth]{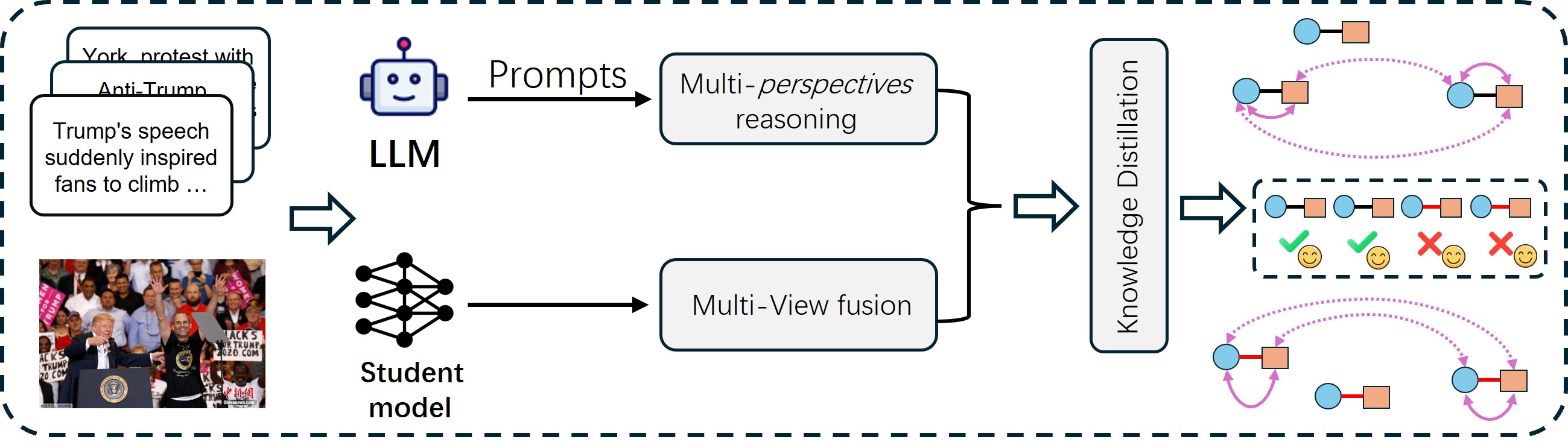}
        \caption{Our method}
        \label{fig:sub_weight_and_heads}
    \end{subfigure}
    \caption{Comparison of our proposed "teacher-student" framework with previous methods. 
(a) Traditional fusion methods (e.g., Attention, Add) lack deep contextual understanding and are easily misled by manipulations like image-text mismatching, leading to misjudgments. 
(b) Our approach employs an LLM teacher for deep multi-view reasoning and distills this knowledge to a efficient student model, enabling it to learn robust features for accurate classification.
}
    \label{fig:framework_comparison_vertical}
\end{figure}

\item We designed a Teacher Multi-view Reasoning approach that bridges the semantic gap between teacher-level reasoning and student-level features. Through calibration distillation, our method measures the reasoning discrepancy between teacher and student embeddings, using a distillation loss to minimize errors. This enables the student to inherit complex reasoning-derived insights via self-calibration.

\item We proposed a Calibration Distillation strategy that explicitly models and fuses multi-source evidence. Unlike prior LLM-based methods that depend on pseudo-labeling or naive aggregation, our strategy cohesively integrates textual, visual, and cross-modal reasoning.

\item We validate the effectiveness of \textbf{LLM-MRD} through extensive experiments on three widely used multimodal FND benchmarks across two different languages, where our scheme
shows significant performance gains compared to the state-of-the-art baselines.
\end{enumerate}

\section{RELATED WORK}
\subsection{Multimodal Fake News Detection}
Early unimodal methods \cite{liu2024fakenewsgpt4,jiang2025cross} proved insufficient against multimodal misinformation, prompting a shift toward multimodal detection \cite{zhou2025robustrealiblemultimodalfake}. Recent research focuses on text–visual fusion via attention mechanisms and contrastive learning to identify cross-modal consistencies \cite{qiu2025dsen}.

However, existing fusion methods often interpret images superficially at the pixel or entity level \cite{lu2025dammfnd}. This limitation, combined with isolated textual features, creates “information islands” that result in weak semantic reasoning. To address this, we leverage LLMs for multi-perspective reasoning, generating rich semantic descriptions to dismantle these silos and enhance cross-modal understanding \cite{wei2022chain}.

\subsection{Application of LLM in Fake News Detection}
Multimodal large language models (MLLMs) have become pivotal in detection frameworks \cite{10843779,zhang2025llms} due to their generative and reasoning capabilities \cite{chen2025lvagent}.

LLMs enhance detection by generating explanatory reasoning chains via prompting \cite{yao2024survey} to uncover subtle manipulation clues \cite{liu2024stepwise}. They also synthesize contradictory arguments to train models on semantic consistency \cite{wang2024llm}. Furthermore, LLMs serve as ``teacher'' models in knowledge distillation \cite{teo2024integrating,tong2024mmdfnd}. The paradigm is shifting from merely imitating outputs to distilling the semantic knowledge of the reasoning process \cite{nam2024using}, enabling student models to capture nuanced cues \cite{nathanson2024step}.

Despite challenges such as hallucinations \cite{huang2025can}, employing LLMs as external reasoning engines \cite{chalehchaleh2025addressing} to transfer capabilities to specialized models remains a robust strategy.

\section{Methodology}
To address multimodal fake news detection, we propose \textbf{LLM-MRD}, LLM-Guided Multi-View Reasoning Distillation for Fake News Detection. Our method distills reasoning knowledge from a powerful MLLM teacher into a efficient student model. As depicted in Figure \ref{fig:model_architecture}, the framework comprises four key components: (1) \textbf{Student Multi-view Reasoning}, (2) \textbf{Teacher Multi-view Reasoning}, (3) \textbf{Calibration Distillation}, and (4) \textbf{Multi-View Fusion} for final prediction.

\subsection{Student Multi-view Reasoning}
The student model acquires robust representations by analyzing news content from three distinct perspectives: textual, visual, and cross-modal.

\indent\textbf{Text View:} For text $T$, a pre-trained BERT\cite{devlin2019bert} extracts contextualized token embeddings. A self-attention mechanism captures salient intra-modal semantics and dependencies. The final text view representation, $f_{\text{text}}$, is:
\begin{equation}
f_{\text{text}} = \text{Self-Attention}(\text{BERT}(T)) \in \mathbb{R}^{d},
\end{equation}

\begin{figure}[htb]
\centering
\includegraphics[width=\textwidth]{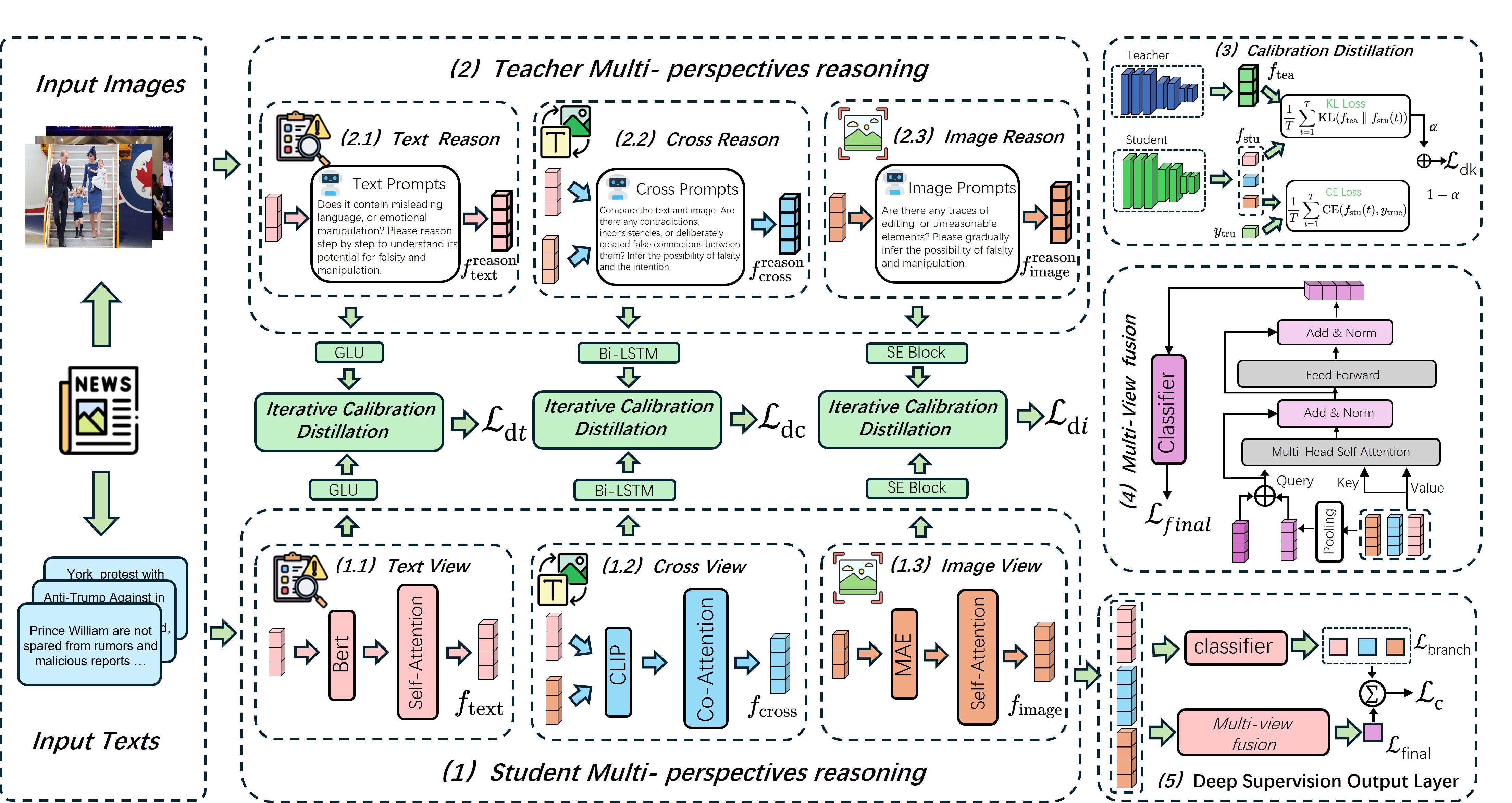}
\caption{LLM-MRD architecture overview. The student uses \textbf{BERT}\cite{devlin2019bert}, \textbf{MAE}\cite{he2022masked}, and \textbf{CLIP}\cite{radford2021learning} to encode textual, visual, and cross-modal view features. The LLM teacher generates multi-view reasoning chains. \textbf{Calibration Distillation} aligns student features with the teacher's semantic space using projection layers before applying distillation losses. Calibrated features are integrated by Multi-View Fusion via cross-attention for prediction, while a Deep Supervision layer provides auxiliary guidance.}
\label{fig:model_architecture}
\end{figure}

where $d$ is the dimension of the feature vector.

\indent\textbf{Image View:} For image $I$, a Masked Autoencoder (MAE)\cite{he2022masked} obtains patch-level visual embeddings. A self-attention layer models spatial dependencies and aggregates global visual context, yielding $f_{\text{image}}$:
\begin{equation}
f_{\text{image}} = \text{Self-Attention}(\text{MAE}(I)) \in \mathbb{R}^{d},
\end{equation}

\indent\textbf{Cross View:} To model alignment and consistency, we use a cross-view perspective. We leverage CLIP\cite{radford2021learning} for aligned image ($f_{\text{CLIP-I}}$) and text ($f_{\text{CLIP-T}}$) embeddings. A co-attention mechanism captures intricate correlations and inconsistencies. The resulting cross-modal representation, $f_{\text{cross}}$, is:
\begin{equation}
f_{\text{cross}} = \text{Co-Attention}(f_{\text{CLIP-I}}, f_{\text{CLIP-T}}) \in \mathbb{R}^{d},
\end{equation}

\subsection{Teacher Multi-view Reasoning}
The teacher model, a powerful MLLM (Qwen2.5-VL), generates high-level, human-like reasoning as rich supervisory signals. Instead of a veracity label, the teacher produces detailed textual explanations from the same three perspectives as the student, guided by prompts.

\indent\textbf{Text Reason:} We prompt the teacher to analyze the text for manipulation. Given a specific prompt $P_{\text{text}}$ and input T, the teacher generates a detailed textual reasoning response, $f^{\text{reason}}_{\text{text}}$.

\indent\textbf{Image Reason:} We prompt the teacher to inspect the image for artifacts or flaws using $P_{\text{image}}$. The teacher generates the visual reasoning response $f^{\text{reason}}_{\text{image}}$ based on image I and $P_{\text{image}}$.

\indent\textbf{Cross Reason:} To evaluate text-image consistency, we use $P_{\text{cross}}$. The teacher takes T, I, and $P_{\text{cross}}$ as input to produce the reasoning $f^{\text{reason}}_{\text{cross}}$. The complete details and templates for all prompts ($P_{\text{text}}$, $P_{\text{image}}$, $P_{\text{cross}}$) are provided in our anonymous supplementary material.\footnote{\raggedright Anonymous code and prompts: \url{https://anonymous.4open.science/r/LLM-MRD-CC50}}

These generated reasoning chains ($f^{\text{reason}}_{\text{text}}$, $f^{\text{reason}}_{\text{image}}$, $f^{\text{reason}}_{\text{cross}}$) are then embedded to distill their high-level, abstract cognition into a structured, dense vector space. We employ a pre-trained Sentence Transformer model \cite{reimers2019sentencebert} for this encoding process. This yields the high-fidelity feature vectors ($f'_{\text{text}}$, $f'_{\text{image}}$, $f'_{\text{cross}}$), which serve as the rich supervisory signals.

\subsection{Calibration Distillation}
To transfer the knowledge from the teacher's multi-perspective \textbf{reasoning}, rather than just mimicking feature outputs, we devise the \textbf{Calibration Distillation} (CD) module.
A naive distillation that directly minimizes the distance between the student's nascent features ($f_v$) and the teacher's reasoning vectors ($f'_v$) via a simple projection layer (e.g., $\hat{f}_v = \text{MLP}_v(f_v)$) is a standard technique. However, this approach merely projects features and fails to transfer the complex reasoning logic itself \cite{nam2024using}.

\indent Our approach re-frames this problem as one of \textbf{self-reflection and correction}. Instead of just projecting $f_v$, we explicitly task the student model with \textit{predicting} a "correction vector" $d_v^{\text{pred}}$ that represents the gap between its own understanding and the teacher's. Crucially, this prediction is informed by a holistic, multi-view context. The module's input is the concatenation of all three student-view features: $F_{\text{concat}} = \text{concat}(f_{\text{text}}, f_{\text{image}}, f_{\text{cross}})$. View-specific MLPs then predict the correction vector for each perspective:
\begin{equation}
d_v^{\text{pred}} = \text{MLP}_v(F_{\text{concat}}) \label{eq:predicted_discrepancy},
\end{equation}

\indent The student's original feature $f_v$ is then \textit{calibrated} by applying this predicted correction in an additive, residual manner. This produces the final calibrated feature $\hat{f}_v$:
\begin{equation}
\hat{f}_v = f_v + d_v^{\text{pred}} \label{eq:calibration_apply},
\end{equation}
This mechanism is fundamentally different from a simple projection. The "calibration" is an additive residual ($+ d_v^{\text{pred}}$) informed by all modalities, directly enabling the student to self-correct its reasoning based on a comprehensive understanding of the input.

\indent Finally, the distillation loss $\mathcal{L}_{dv}$ is designed to optimize this self-correction
process. It trains the MLP (Eq. \ref{eq:predicted_discrepancy}) by minimizing the discrepancy between the
teacher's target reasoning $f'_v$ and the student's calibrated feature $\hat{f}_v$. This loss combines two objectives:

\indent\textbf{Knowledge Distillation Loss (KL Divergence).} The primary loss, KL
divergence, minimizes the distributional difference between the teacher's features
and the student's calibrated features. As $f'_{v}$ (teacher) and $\hat{f}_v$ (calibrated student)
are feature vectors, we first convert them into probability distributions using a
temperature-scaled softmax function ($\sigma$). The temperature $\tau$, analyzed in Sec.
4.4, controls the softness of these distributions.

\indent\textbf{Auxiliary Task Loss (Cross-Entropy).} Concurrently, we apply an auxiliary
cross-entropy (CE) loss to each \textit{individual} calibrated feature $\hat{f}_v$.
This ensures that each view-specific representation $\hat{f}_v$ is independently
discriminative and predictive of the ground-truth label $y_{\text{true}}$ \textit{before}
they are integrated by the fusion module.

\noindent This combined loss $\mathcal{L}_{dv}$ is formulated as:
\begin{equation}
\mathcal{L}_{dv} = \alpha \cdot (\tau^2) \cdot \text{KL}(\sigma(f'_v, \tau) \Vert \sigma(\hat{f}_v, \tau)) + (1 - \alpha) \cdot \text{CE}(\hat{f}_v, y_{\text{true}}), \label{eq:distill_loss} 
\end{equation}
where $\alpha$ is a weighting coefficient.
This loss serves a dual purpose: the KL term aligns the student with the teacher's
reasoning space, while the CE term acts as a \textbf{deep supervision signal} for each modality.
This design prevents any single view from becoming "lost" during optimization,
ensuring all views contribute meaningfully to the final fused prediction, which is
separately supervised by $\mathcal{L}_C$ (Sec 3.5).

These calibrated features $\hat{f}_v$ are then passed to the fusion module.

\subsection{Multi-View Fusion}
The Multi-View Fusion module's distinction is its core attention. While standard Transformers use self-attention, our module is purpose-built around cross-attention for heterogeneous feature fusion.

\indent\textbf{The Query (High-Level Summary):} The query is $\hat{F}_{\text{pool}}$, a single, aggregated feature vector produced by pooling all calibrated perspectives ($\hat{f}_{\text{text}}$, $\hat{f}_{\text{image}}$, $\hat{f}_{\text{cross}}$) from Section 3.3. It represents a holistic, self-corrected summary of the content.
\begin{equation}
\text{Attention}(Q, K, V) = \text{softmax}\left(\frac{Q K^T}{\sqrt{d_k}}\right) V,
\end{equation}

The cross-attention mechanism facilitates a dynamic interaction where the summary query ($\hat{F}_{\text{pool}}$) actively probes this detailed evidence ($K=\hat{F}_{\text{views}}, V=\hat{F}_{\text{views}}$). It learns to identify and dynamically up-weight the most relevant modal features relative to the overall context, producing a deeply integrated and context-aware fused representation, $F_{\text{final}}$:
\begin{equation}
F_{\text{final}} = \text{CrossAttention}(Q=\hat{F}_{\text{pool}}, K=\hat{F}_{\text{views}}, V=\hat{F}_{\text{views}}),
\end{equation}

\subsection{Loss Function}
\textbf{Classification Loss ($\mathcal{L}_C$).} This component ensures task-specific accuracy using deep supervision. It combines two Cross-Entropy (CE) losses: (1) The main loss, $\mathcal{L}_{\textit{final}}$, computed from the final fused feature $F_{\textit{final}}$, and (2) an auxiliary branch loss, $\mathcal{L}_{\textit{branch}}$, computed from the uncalibrated student features ($f_{\text{text}}$, $f_{\text{image}}$, $f_{\text{cross}}$) to guide early-stage learning. $\mathcal{L}_C$ is the sum of these two:
\begin{align}
\mathcal{L}_{\textit{final}} &= \text{CE}(\text{Classifier}(F_{\textit{final}}), y), \\
\mathcal{L}_C &= \mathcal{L}_{\textit{final}} + \mathcal{L}_{\textit{branch}},
\end{align}

\indent The overall training objective, $\mathcal{L}_{total}$, is a weighted combination of the main classification loss $\mathcal{L}_C$ and the view-specific calibration losses $\mathcal{L}_{dv}$ introduced in Section 3.3 (Eq. \ref{eq:distill_loss}). The total loss is defined as:
\begin{equation}
\mathcal{L}_{total} = \mathcal{L}_C + \lambda \sum_{v \in \{\text{text}, \text{cross}, \text{image}\}} \mathcal{L}_{dv}.
\end{equation}
where $\lambda$ is a balance coefficient.

\section{Experiments}
\subsection{Experimental Settings}
We conduct extensive experiments to validate LLM-MRD, detailing our datasets, baselines, and implementation.

\indent\textbf{Datasets.}
We evaluate on three standard public benchmarks: \textbf{Weibo}\cite{wang2018eann}, \textbf{Weibo21}\cite{zhou2020safe}, and \textbf{GossipCop}\cite{liu2025modality}, using well-established partitioning and pre-processing. \textbf{Weibo}\cite{wang2018eann} has 7,532 training (3,749 real, 3,783 fake) and 1,996 testing (996 real, 1,000 fake) articles. \textbf{Weibo21}\cite{zhou2020safe} contains 9,127 articles (4,640 real, 4,487 fake). \textbf{GossipCop}\cite{liu2025modality} has 10,010 training (7,974 real, 2,036 fake) and 2,830 testing (2,285 real, 545 fake) articles.

\indent\textbf{Baselines.}
We benchmark against three baseline categories: (1) \textit{Unimodal methods} (MVAN\cite{ni2021mvan}, SpotFake\cite{singhal2019spotfake}). (2) Cross-domain generalization methods (including EANN\cite{wang2018eann}, FND-CLIP\cite{zhou2023multimodal}, MIMoE-FND\cite{liu2025modality},KEN\cite{zhu2025ken}). (3) LLM distillation methods (GLPN-LLM\cite{hu2025synergizing}, FactAgent\cite{li2024large}, LLM-GAN\cite{wang2024llm}).

\indent\textbf{Implementation Details.}
We use "mae-pretrain-vit-base"\cite{he2022masked} as the image encoder (images 224 × 224). Text encoders are "bert-base-chinese"\cite{devlin2019bert} for Weibo/Weibo21 and "bert-base-uncased"\cite{devlin2019bert} for GossipCop, truncated to 197 tokens. CLIP embeddings use the "clip-vit-base-patch16"\cite{radford2021learning} (GossipCop) and "chinese-clip-vit-base-patch16"\cite{yang2022chinese}. The teacher model is Qwen2.5-VL\cite{liu2025modality} with 12 attention heads. Experiments use PyTorch on one NVIDIA RTX 4090 GPU.

\indent\textbf{The Context (Detailed Multi-View Evidence):} In contrast, the Key and Value are derived from the set of individual, un-pooled calibrated perspective features ($\hat{f}_{\text{text}}$, $\hat{f}_{\text{image}}$, $\hat{f}_{\text{cross}}$), which we denote as $\hat{F}_{\text{views}}$. This set acts as a detailed knowledge repository, preserving the distinct and nuanced information from each calibrated modality:
\begin{equation}
\hat{F}_{\text{views}} = \{\hat{f}_{\text{text}}, \hat{f}_{\text{image}}, \hat{f}_{\text{cross}}\}
\end{equation}

\begin{table}[htbp] 
    \centering
    \caption{Comparison of LLM-MRD with the latest multi-domain fake news detection methods from Weibo, Weibo-21, and GossipCop.}
    \label{tab:main_results}
    \resizebox{\textwidth}{!}{
    \begin{tabular}{llrrrrrrrrrrrr}
        \toprule 
         \rowcolor{lightcyan} 
        & & \multicolumn{4}{>{\columncolor{lightcyan}}c}{\textbf{Weibo}} & \multicolumn{4}{>{\columncolor{lightcyan}}c}{\textbf{Weibo-21}} & \multicolumn{4}{>{\columncolor{lightcyan}}c}{\textbf{GossipCop}} \\
         \cmidrule(lr){3-6} \cmidrule(lr){7-10} \cmidrule(lr){11-14} 
         \rowcolor{lightcyan}
         \hspace{2.5em}\textbf{Category} & \textbf{Method} & \textbf{Acc} & \textbf{F1-Fake} & \textbf{F1-Real} & \textbf{AUC} & \textbf{Acc} & \textbf{F1-Fake} & \textbf{F1-Real} & \textbf{AUC}  & \textbf{Acc} & \textbf{F1-Fake} & \textbf{F1-Real} & \textbf{AUC}  \\
        \midrule 
         \multirow{4}{*}{\shortstack{Multimodal \\ multi-domain methods}}
         & EANN\cite{wang2018eann} & 0.827 & 0.829 & 0.825 & 0.873 & 0.870 & 0.862 & 0.875 & 0.894 & 0.864 & 0.594 & 0.920 & 0.852\\
         & FND-CLIP\cite{zhou2023multimodal} & 0.907 & 0.908 & 0.907 & 0.953 & 0.943 & 0.940 & 0.946 & 0.962 & 0.880 & 0.638 & 0.928 & 0.871\\
         & MIMoE-FND\cite{liu2025modality} & 0.928 & 0.928 & 0.928  & 0.972 & 0.956 & 0.955 & 0.957 & 0.977 & 0.895 & 0.698 & 0.938 & 0.879\\
         & KEN\cite{zhu2025ken} & 0.935 & 0.935 & 0.934 & 0.967 & 0.935 & 0.937 & 0.932 & 0.971 & 0.881 & 0.646 & 0.928 & 0.873 \\
        \midrule 
         \multirow{5}{*}{\shortstack{Multimodal \\ single-domain methods}}
         & RaCMC\cite{yu2025racmc} & 0.915 & 0.917 & 0.914 & 0.921 & 0.942 & 0.938 & 0.943 & 0.962 & 0.879 & 0.641 & 0.927 & 0.838 \\
         & SAFE\cite{zhou2020safe} & 0.762 & 0.774 & 0.748 & 0.824 & 0.905 & 0.901 & 0.890 & 0.937 & 0.838 & 0.643 & 0.895 & 0.824 \\
         & CAFE\cite{chen2022cross} & 0.840 & 0.842 & 0.837 & 0.892 & 0.882 & 0.885 & 0.876 & 0.909 & 0.867 & 0.587 & 0.921 & 0.852 \\
         & BMR\cite{ying2023bootstrapping} & 0.918 & 0.914 & 0.904 & 0.954 & 0.929 & 0.927 & 0.925 & 0.962 & 0.895 & 0.691 & 0.876 & 0.881 \\
         & SEER\cite{zhu2025seer} & 0.929& 0.928 & 0.939 & 0.934 & 0.932 & 0.927 & 0.925 & 0.960 & 0.893 & 0.673 & 0.871 & 0.875 \\
        \midrule 
         \multirow{3}{*}{\shortstack{Detection methods \\ that distill LLMs}}
         & LLM-GAN\cite{wang2024llm} & 0.813 & 0.822 & 0.819 & 0.864 & 0.806 & 0.796 & 0.812 & 0.851 & 0.896 & 0.712 & 0.941 & 0.877\\
         & GLPN-LLM\cite{hu2025synergizing} & 0.920 & 0.939 & 0.921 & 0.954 & 0.925 & 0.937 & 0.924 & 0.959 & 0.890 & 0.682 & 0.933 & 0.864\\
         & FactAgent\cite{li2024large} & 0.927 & 0.943 & 0.925 & 0.962 & 0.932 & 0.938 & 0.936 & 0.973 & 0.860 & 0.854 & 0.862 & 0.851\\
        \midrule 
         \multirowcell{3}{\textbf{Ours}}
         & \textbf{LLM-MRD} & \textcolor{red}{0.953} & \textcolor{red}{0.954} & \textcolor{red}{0.952} & \textcolor{red}{0.987} & \textcolor{red}{0.959} & \textcolor{red}{0.957} & \textcolor{red}{0.968} & \textcolor{red}{0.988} & \textcolor{red}{0.902} & \textcolor{red}{0.719} & \textcolor{red}{0.946} & \textcolor{red}{0.885} \\
         & Improv. & 1.93\%$\uparrow$ & 1.17\%$\uparrow$ & 1.38\%$\uparrow$ & 1.54\%$\uparrow$ & 0.31\%$\uparrow$ & 0.21\%$\uparrow$ & 1.15\%$\uparrow$ & 1.13\%$\uparrow$ & 0.67\%$\uparrow$ & 0.98\%$\uparrow$ & 0.53\%$\uparrow$ & 0.45\%$\uparrow$ \\
         & $p$-val. & $9.72e^{-3}$ & $8.52e^{-3}$ & $7.44e^{-3}$ & $7.51e^{-3}$ & $9.91e^{-4}$ & $3.07e^{-4}$ & $1.47e^{-3}$ & $3.67e^{-4}$ & $2.29e^{-4}$ & $1.68e^{-3}$ & $2.62e^{-4}$ & $3.61e^{-3}$ \\
        \bottomrule 
    \end{tabular}%
    }
\end{table}

\subsection{Overall Performance}
To validate LLM-MRD's superiority, we compare it against 12 competitive baselines on three datasets (Table \ref{tab:main_results}). We draw the following observations:

\indent\textbf{(O1): Multi-domain methods generally outperform their single-domain counterparts.}
Single-domain methods, trained on specific data, suffer from poor generalization and ignore feature discrepancies, resulting in weaker performance. Multi-domain methods enhance cross-domain accuracy by leveraging shared knowledge. LLM-MRD performs best, as competing approaches often use naive parameter sharing that disrupts domain patterns and lack specialized designs to digest high-level LLM reasoning, thus limiting their generalization.

\indent\textbf{(O2): Our framework, based on calibration distillation, demonstrates outstanding performance.}
Table \ref{tab:main_results} shows LLM-MRD consistently outperforms competing methods. The key is our Calibration Distillation module. Rather than superficial knowledge transfer, it distills the rich semantic output of the teacher LLM’s multi-view reasoning process. The student progressively aligns its representations with the teacher’s semantic space, inheriting sophisticated reasoning-derived knowledge. This yields a more robust model with stronger cross-modal understanding, explaining its superior performance.

\indent\textbf{(O3): Our LLM reasoning distillation paradigm is superior to other LLM-based strategies.}
Among LLM-based methods, LLM-MRD shows a clear advantage. Competing methods such as LLM-GAN\cite{wang2024llm} or GLPN-LLM\cite{hu2025synergizing} use LLMs for adversarial training or feature enhancement but fail to transfer the full nuanced reasoning process. In contrast, our framework explicitly distills the semantic conclusions of how the teacher reasons across modalities, yielding deeper understanding of inconsistencies and a significant performance margin (Table \ref{tab:main_results}).

\indent\textbf{(O4): LLM-MRD outperforms all strong baselines across all three datasets.}
Notably, LLM-MRD demonstrates a comprehensive performance lead. When averaged across all 12 baseline methods and all three datasets, our model achieves an average relative improvement of 5.19\% in ACC, 6.33\% in F1-Fake, and 5.63\% in F1-Real. To confirm this superiority against the strongest baseline, we retrained LLM-MRD and the best baseline five times. All resulting $p$-values are below 0.05, indicating statistically significant improvements. 
These results demonstrate the effectiveness of our large model inference and calibration distillation.

\begin{table}[t]
\centering
\caption{Ablation study results showing performance drops for different model variants.}
\label{tab:ablation_moda_llm}

\resizebox{\linewidth}{!}{%
\begin{tabular}{lccccccccc}
\toprule
\rowcolor{lightcyan}
\multirow{2}{*}{Model} & \multicolumn{3}{c}{\textbf{Weibo}} & \multicolumn{3}{c}{\textbf{Weibo-21}} & \multicolumn{3}{c}{\textbf{GossipCop}} \\
\cmidrule(lr){2-4} \cmidrule(lr){5-7} \cmidrule(lr){8-10}
\rowcolor{lightcyan}
& \textbf{Acc} & \textbf{F1-Fake} & \textbf{F1-Real} & \textbf{Acc} & \textbf{F1-Fake} & \textbf{F1-Real} & \textbf{Acc} & \textbf{F1-Fake} & \textbf{F1-Real} \\
\midrule
\textbf{LLM-MRD} & 0.953 & 0.954 & 0.952 & 0.959 & 0.957 & 0.968 & 0.902 & 0.719 & 0.946 \\
\midrule
\multicolumn{10}{l}{\textit{Iterative Calibration distillation}} \\
- w/o $\mathcal{L}_{\text{text}}$ & 0.929\,\textcolor{red!80!black}{\tiny{$\downarrow$2.5\%}} & 0.924\,\textcolor{red!80!black}{\tiny{$\downarrow$3.1\%}} & 0.920\,\textcolor{red!80!black}{\tiny{$\downarrow$3.4\%}} & 0.937\,\textcolor{red!80!black}{\tiny{$\downarrow$2.3\%}} & 0.933\,\textcolor{red!80!black}{\tiny{$\downarrow$2.5\%}} & 0.938\,\textcolor{red!80!black}{\tiny{$\downarrow$3.1\%}} & 0.881\,\textcolor{red!80!black}{\tiny{$\downarrow$2.3\%}} & 0.672\,\textcolor{red!80!black}{\tiny{$\downarrow$6.5\%}} & 0.915\,\textcolor{red!80!black}{\tiny{$\downarrow$3.3\%}} \\
- w/o $\mathcal{L}_{\text{image}}$ & 0.925\,\textcolor{red!80!black}{\tiny{$\downarrow$2.9\%}} & 0.918\,\textcolor{red!80!black}{\tiny{$\downarrow$3.8\%}} & 0.922\,\textcolor{red!80!black}{\tiny{$\downarrow$3.2\%}} & 0.923\,\textcolor{red!80!black}{\tiny{$\downarrow$3.8\%}} & 0.935\,\textcolor{red!80!black}{\tiny{$\downarrow$2.3\%}} & 0.937\,\textcolor{red!80!black}{\tiny{$\downarrow$3.2\%}} & 0.874\,\textcolor{red!80!black}{\tiny{$\downarrow$3.1\%}} & 0.674\,\textcolor{red!80!black}{\tiny{$\downarrow$6.3\%}} & 0.913\,\textcolor{red!80!black}{\tiny{$\downarrow$3.5\%}} \\
- w/o $\mathcal{L}_{\text{cross}}$ & 0.917\,\textcolor{red!80!black}{\tiny{$\downarrow$3.8\%}} & 0.907\,\textcolor{red!80!black}{\tiny{$\downarrow$4.9\%}} & 0.912\,\textcolor{red!80!black}{\tiny{$\downarrow$4.2\%}} & 0.945\,\textcolor{red!80!black}{\tiny{$\downarrow$1.5\%}} & 0.921\,\textcolor{red!80!black}{\tiny{$\downarrow$3.8\%}} & 0.927\,\textcolor{red!80!black}{\tiny{$\downarrow$4.2\%}} & 0.889\,\textcolor{red!80!black}{\tiny{$\downarrow$1.4\%}} & 0.665\,\textcolor{red!80!black}{\tiny{$\downarrow$7.5\%}} & 0.901\,\textcolor{red!80!black}{\tiny{$\downarrow$4.8\%}} \\
\midrule
\multicolumn{10}{l}{\textit{Multi-View fusion}} \\
- w/o Text View & 0.935\,\textcolor{red!80!black}{\tiny{$\downarrow$1.9\%}} & 0.933\,\textcolor{red!80!black}{\tiny{$\downarrow$2.2\%}} & 0.928\,\textcolor{red!80!black}{\tiny{$\downarrow$2.5\%}} & 0.948\,\textcolor{red!80!black}{\tiny{$\downarrow$1.1\%}} & 0.939\,\textcolor{red!80!black}{\tiny{$\downarrow$1.9\%}} & 0.943\,\textcolor{red!80!black}{\tiny{$\downarrow$2.6\%}} & 0.884\,\textcolor{red!80!black}{\tiny{$\downarrow$2.0\%}} & 0.686\,\textcolor{red!80!black}{\tiny{$\downarrow$4.6\%}} & 0.923\,\textcolor{red!80!black}{\tiny{$\downarrow$2.4\%}} \\
- w/o Image View & 0.936\,\textcolor{red!80!black}{\tiny{$\downarrow$1.8\%}} & 0.927\,\textcolor{red!80!black}{\tiny{$\downarrow$2.8\%}} & 0.924\,\textcolor{red!80!black}{\tiny{$\downarrow$2.9\%}} & 0.952\,\textcolor{red!80!black}{\tiny{$\downarrow$0.7\%}} & 0.936\,\textcolor{red!80!black}{\tiny{$\downarrow$2.2\%}} & 0.939\,\textcolor{red!80!black}{\tiny{$\downarrow$3.0\%}} & 0.874\,\textcolor{red!80!black}{\tiny{$\downarrow$3.1\%}} & 0.682\,\textcolor{red!80!black}{\tiny{$\downarrow$5.1\%}} & 0.915\,\textcolor{red!80!black}{\tiny{$\downarrow$3.3\%}} \\
\midrule
\multicolumn{10}{l}{\textit{Teacher reasoning}} \\
- w/o Reasoning prompts & 0.918\,\textcolor{red!80!black}{\tiny{$\downarrow$3.7\%}} & 0.920\,\textcolor{red!80!black}{\tiny{$\downarrow$3.6\%}} & 0.922\,\textcolor{red!80!black}{\tiny{$\downarrow$3.2\%}} & 0.913\,\textcolor{red!80!black}{\tiny{$\downarrow$4.8\%}} & 0.926\,\textcolor{red!80!black}{\tiny{$\downarrow$3.2\%}} & 0.928\,\textcolor{red!80!black}{\tiny{$\downarrow$4.1\%}} & 0.858\,\textcolor{red!80!black}{\tiny{$\downarrow$4.9\%}} & 0.683\,\textcolor{red!80!black}{\tiny{$\downarrow$5.0\%}} & 0.921\,\textcolor{red!80!black}{\tiny{$\downarrow$2.6\%}} \\
- w/o LLM & 0.902\,\textcolor{red!80!black}{\tiny{$\downarrow$5.4\%}} & 0.916\,\textcolor{red!80!black}{\tiny{$\downarrow$4.0\%}} & 0.914\,\textcolor{red!80!black}{\tiny{$\downarrow$4.0\%}} & 0.921\,\textcolor{red!80!black}{\tiny{$\downarrow$4.0\%}} & 0.931\,\textcolor{red!80!black}{\tiny{$\downarrow$2.7\%}} & 0.936\,\textcolor{red!80!black}{\tiny{$\downarrow$3.3\%}} & 0.854\,\textcolor{red!80!black}{\tiny{$\downarrow$5.3\%}} & 0.680\,\textcolor{red!80!black}{\tiny{$\downarrow$5.4\%}} & 0.918\,\textcolor{red!80!black}{\tiny{$\downarrow$3.0\%}} \\
\midrule
\multicolumn{10}{l}{\textit{Student Multi-inputs}} \\
- w/o Feature Extractors & 0.921\,\textcolor{red!80!black}{\tiny{$\downarrow$3.4\%}} & 0.926\,\textcolor{red!80!black}{\tiny{$\downarrow$2.9\%}} & 0.930\,\textcolor{red!80!black}{\tiny{$\downarrow$2.3\%}} & 0.923\,\textcolor{red!80!black}{\tiny{$\downarrow$3.8\%}} & 0.933\,\textcolor{red!80!black}{\tiny{$\downarrow$2.5\%}} & 0.937\,\textcolor{red!80!black}{\tiny{$\downarrow$3.2\%}} & 0.868\,\textcolor{red!80!black}{\tiny{$\downarrow$3.8\%}} & 0.678\,\textcolor{red!80!black}{\tiny{$\downarrow$5.7\%}} & 0.920\,\textcolor{red!80!black}{\tiny{$\downarrow$2.7\%}} \\
- w/o Attention & 0.928\,\textcolor{red!80!black}{\tiny{$\downarrow$2.6\%}} & 0.925\,\textcolor{red!80!black}{\tiny{$\downarrow$3.0\%}} & 0.926\,\textcolor{red!80!black}{\tiny{$\downarrow$2.7\%}} & 0.937\,\textcolor{red!80!black}{\tiny{$\downarrow$2.3\%}} & 0.934\,\textcolor{red!80!black}{\tiny{$\downarrow$2.4\%}} & 0.939\,\textcolor{red!80!black}{\tiny{$\downarrow$3.0\%}} & 0.892\,\textcolor{red!80!black}{\tiny{$\downarrow$1.1\%}} & 0.684\,\textcolor{red!80!black}{\tiny{$\downarrow$4.9\%}} & 0.926\,\textcolor{red!80!black}{\tiny{$\downarrow$2.1\%}} \\
\bottomrule
\end{tabular}
}%
\end{table}

\subsection{Ablation Study}
To understand the role of core components and knowledge distillation in our proposed LLM-MRD framework, we conducted a comprehensive ablation study. Results are summarized in Table \ref{tab:ablation_moda_llm}.

\textbf{Effect of Multi-View Features and Fusion.} We assessed the contribution of individual student views by removing the Text View (\textbf{w/o Text View}) and Image View (\textbf{w/o Image View}) features before fusion. Removing either view, especially the image view, leads to performance degradation, confirming the necessity of integrating information from all three perspectives. This highlights the effectiveness of our multi-view student architecture.

\begin{figure}[htbp]
    \centering

    \begin{subfigure}{0.48\textwidth} 
        \includegraphics[width=\linewidth]{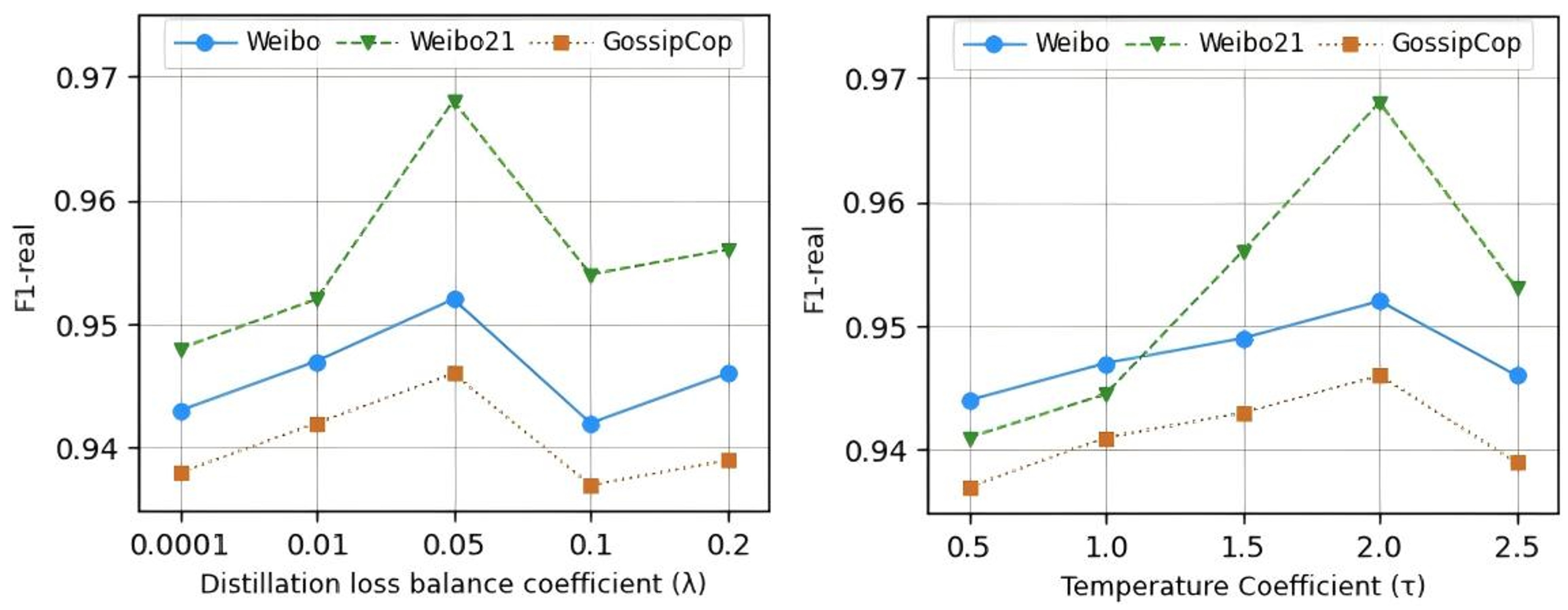}
        \caption{Effects of distillation loss balance coefficient ($\lambda$) and KL divergence temperature coefficients.}
        \label{fig:hyper_a}
    \end{subfigure}
    \hfill 
    \begin{subfigure}{0.48\textwidth} 
        \includegraphics[width=\linewidth]{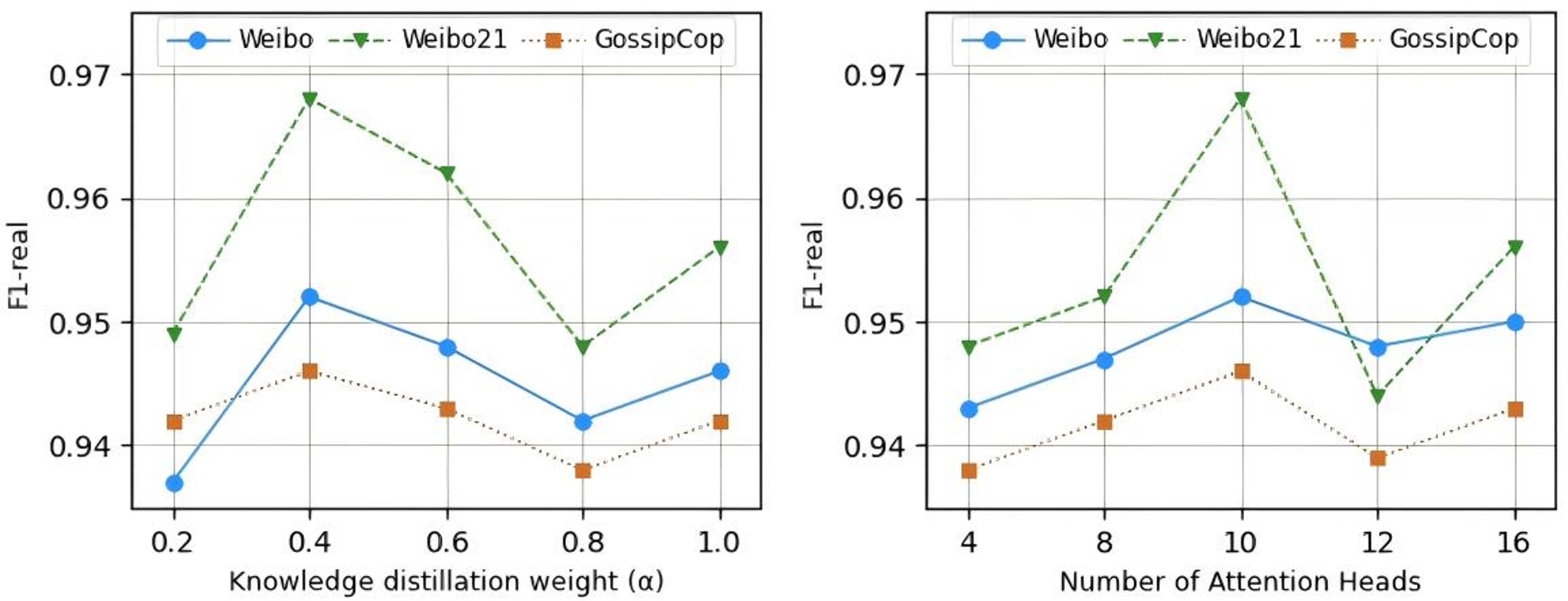}
        \caption{Effects of knowledge distillation weight ($\alpha$) and the number of attention heads.}
        \label{fig:hyper_b}
    \end{subfigure}


    \caption{Analysis of hyperparameter sensitivity. This figure shows the impact of four different hyperparameters on the model's F1-real score across three datasets.}
    \label{fig:hyperparameter_sensitivity}
\end{figure}

\textbf{Effect of Calibration Distillation Components.} We evaluated the impact of distillation from each perspective by removing the specific distillation loss term: without text distillation (\textbf{w/o $\mathcal{L}_{\text{text}}$}), image distillation (\textbf{w/o $\mathcal{L}_{\text{image}}$}), and cross-modal distillation (\textbf{w/o $\mathcal{L}_{\text{cross}}$}). Removing any distillation loss hurts performance, with $\mathcal{L}_{\text{image}}$ showing the largest impact. This verifies the importance of distilling knowledge from all three teacher reasoning perspectives.

\textbf{Effect of Teacher Reasoning and Student Inputs.} Further experiments validate the critical role of the teacher reasoning components. As shown in the table, removing the LLM teacher's guidance (w/o LLM) results in a severe performance drop, with accuracy falling by 5.2\% on Weibo, 3.7\% on Weibo-21, and 4.8\% on GossipCop. Likewise, removing the specialized Reasoning prompts (w/o Reasoning prompts) also causes a significant degradation across all datasets, with accuracy dropping by 3.4\% on Weibo and 4.2\% on Weibo-21. Additionally, removing the student's Feature Extractors or Attention mechanisms leads to notable performance degradation, validating the student's architectural design. Overall, these results demonstrate the effectiveness of each component within the LLM-MRD framework.

\subsection{Parameter sensitivity analysis}
We conducted a comprehensive test of hyperparameter sensitivity on the Weibo, Weibo-21, and GossipCop datasets, focusing on the distillation loss balance $\lambda$, the number of multi-head attention heads, the KL divergence temperature, and the knowledge distillation weight $\alpha$.
As shown in Figure \ref{fig:hyperparameter_sensitivity}, LLM-MRD performs well across various settings. Even in the worst case, it achieves an F1 score of 0.935, demonstrating the model is robust and not overly sensitive.
Figure \ref{fig:hyper_a} shows performance for $\lambda$ forms a bell-shaped curve. Large $\lambda$ values force excessive teacher imitation, neglecting true labels and acquiring biases, while small values fail to fully utilize the teacher's guidance.
From Figure \ref{fig:hyper_b}, we observe performance peaks at 12 attention heads before declining, indicating that both too few and too many heads impair overall performance.

\begin{table}[htbp]
\centering
\vspace{-12pt}
\caption{LLM-MRD explainability case study on two challenging examples from the Weibo21 test set.}
\label{tab:adaptive_structure_transposed}
\resizebox{\linewidth}{!}{%
\renewcommand{\arraystretch}{1.5}%
\begin{tabular}{>{\bfseries}l c c c c c c}
\toprule
& \textbf{\makecell{News\\Posts}} 
& \textbf{\makecell{False\\Type}} 
& \textbf{\makecell{Ground\\Truth}} 
& \textbf{\makecell{Calibration\\Allocation}} 
& \textbf{\makecell{LLM-MRD}} 
& \textbf{\makecell{MIMoE-FND}} \\
\midrule
\raisebox{3\height}{Case 1} &
\includegraphics[width=0.2\textwidth]{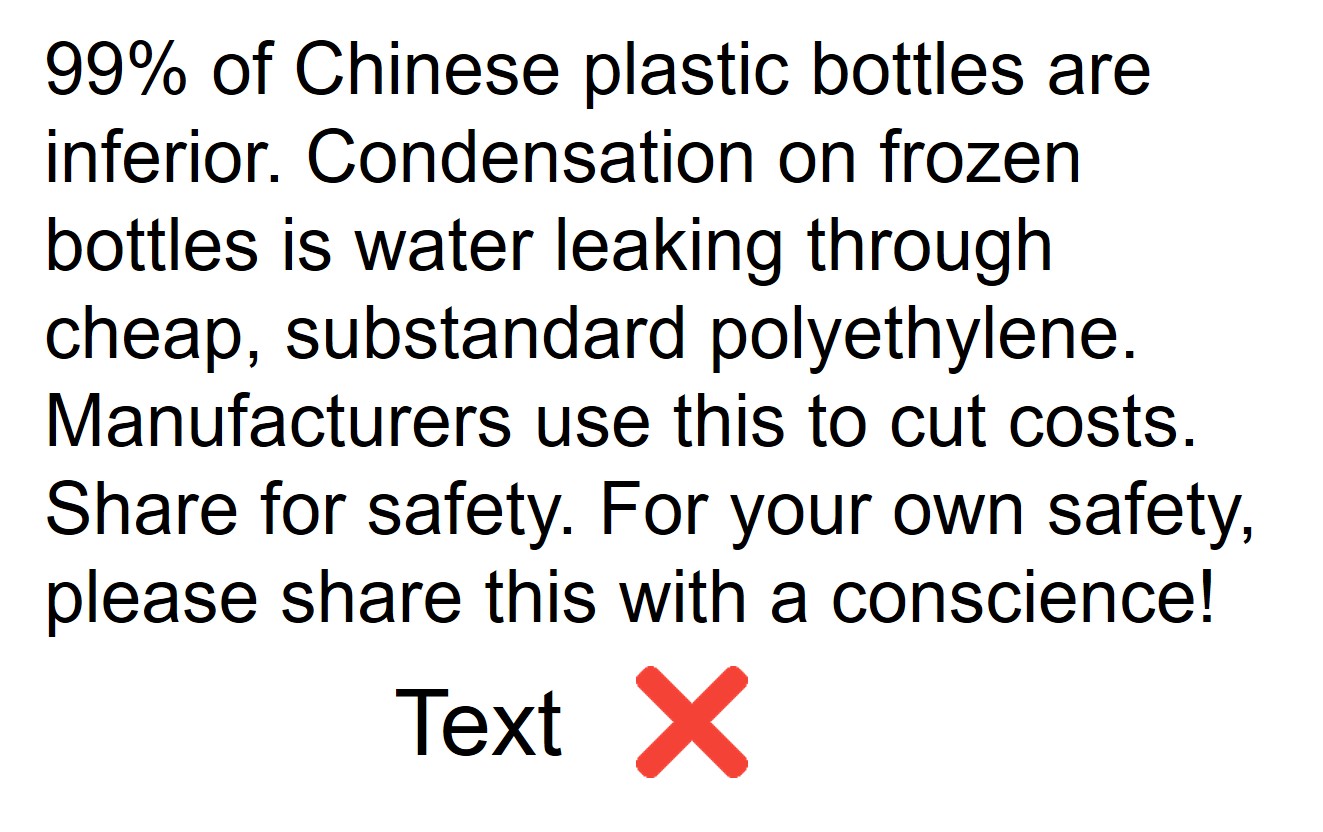} &
\raisebox{1.2\height}{\makecell{Text \\ Fabrication}}&
\raisebox{1.2\height}{\makecell{Fake \\ news}} &
\includegraphics[width=0.2\textwidth]{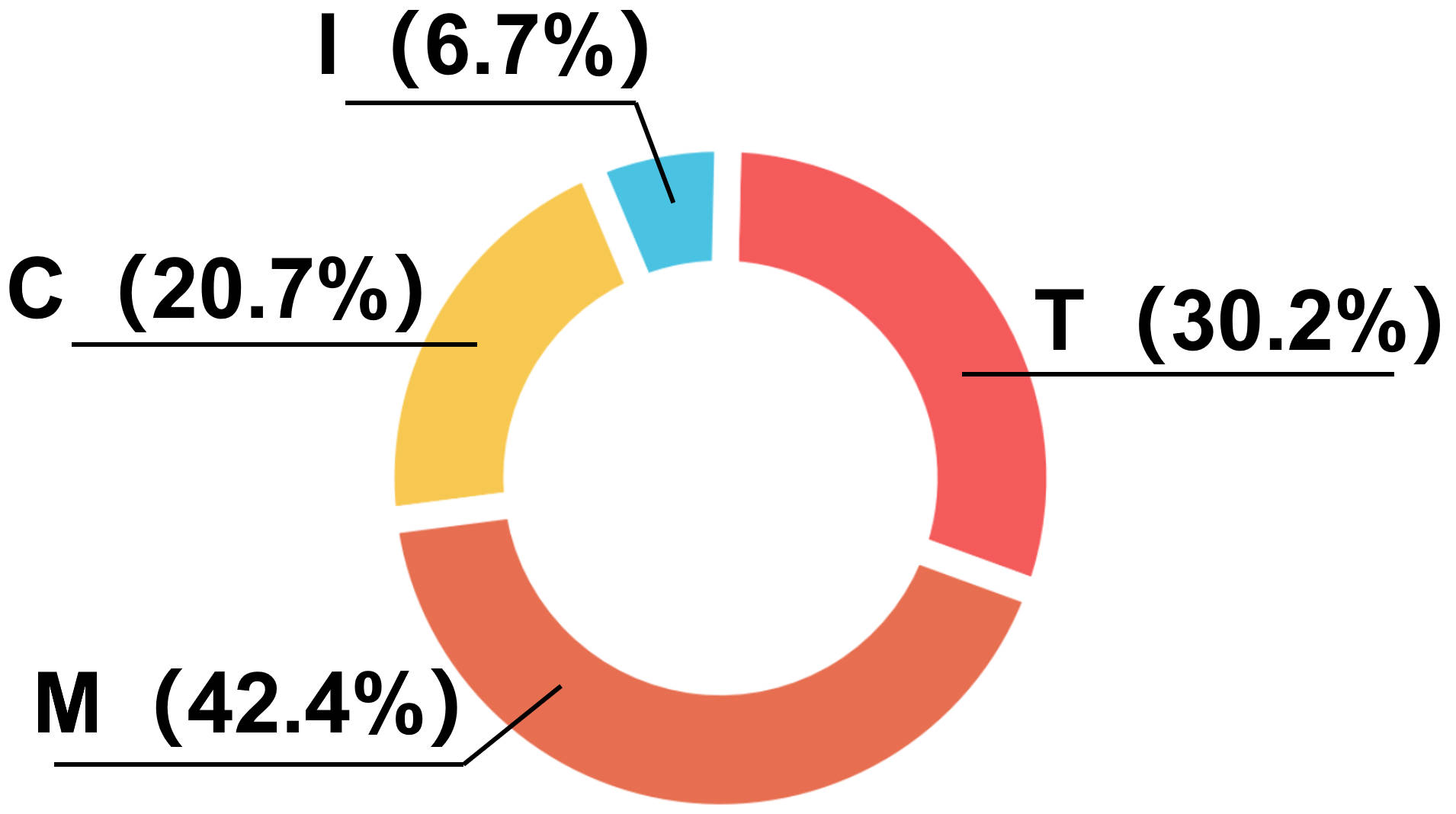} &
\raisebox{1.2\height}{\makecell{Fake \\ news}}  &
\raisebox{1.2\height}{\makecell{Fake \\ news}}  \\
\midrule
\raisebox{3\height}{Case 2} &
\includegraphics[width=0.2\textwidth]{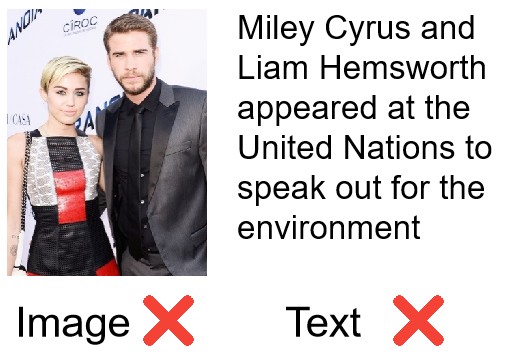} &
\raisebox{1.2\height}{\makecell{Image-Text \\ Mismatch}} &
\raisebox{1.2\height}{\makecell{Fake \\ news}} &
\includegraphics[width=0.2\textwidth]{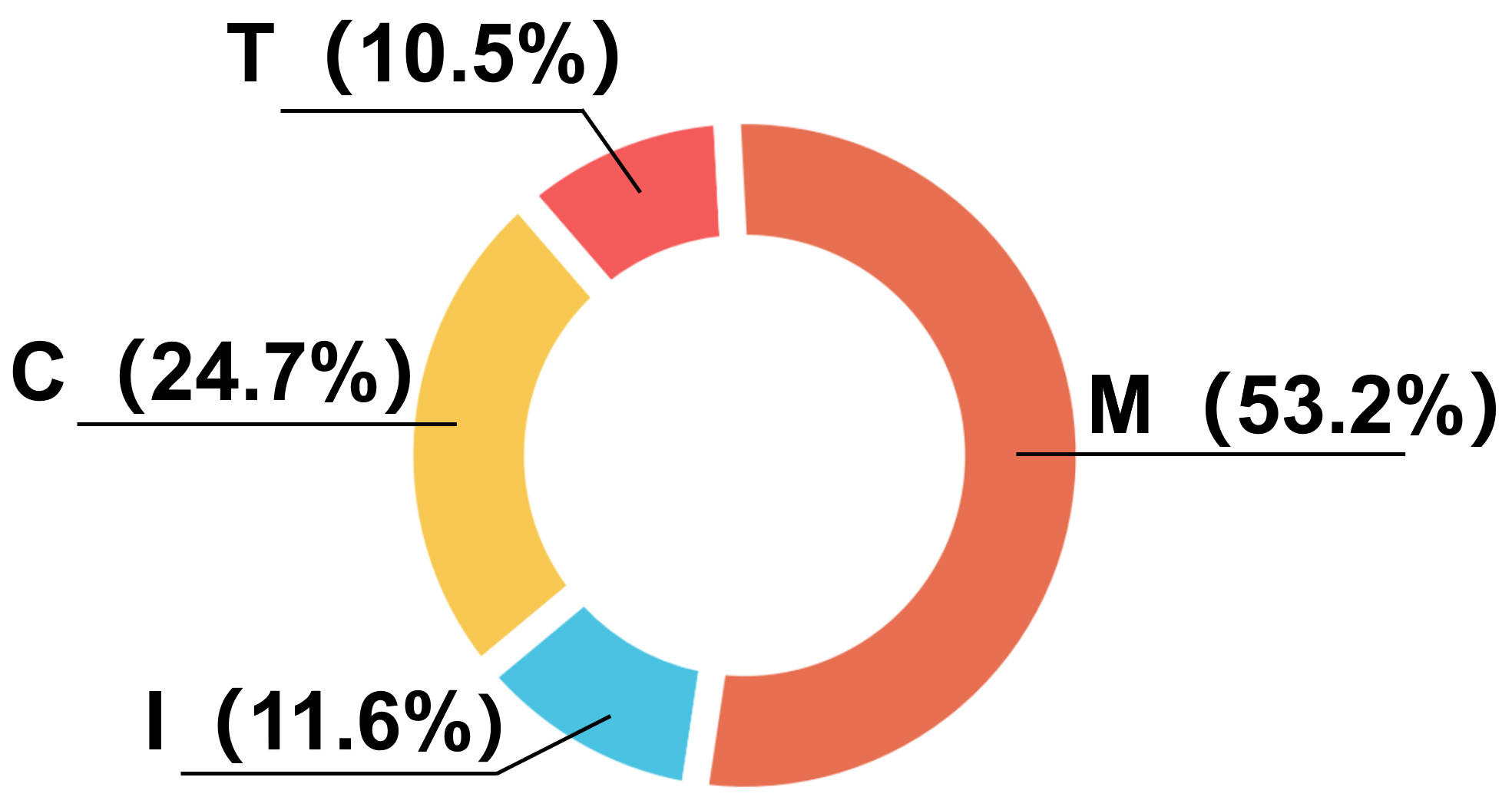} &
\raisebox{1.2\height}{\makecell{Fake \\ news}} &
\raisebox{1.2\height}{\makecell{Real \\ news}}  \\
\bottomrule
\end{tabular}
} 
\vspace{-4pt}
\end{table}

\subsection{Case Study: Model Explainability}
We qualitatively investigate the interpretability of LLM-MRD via a case study of two challenging fake news instances from the Weibo21 test set. This analysis illustrates how LLM-MRD acquires the teacher's deep \textbf{reasoning-derived knowledge} and distills this complex capability into its internal text and image representations.

\indent
The first case involves a text-based fabrication. 
In this instance, both LLM-MRD and the baseline MIMoE-FND\cite{liu2025modality} correctly identified the item as fake news.

\indent
The second case presents a challenging image-text mismatch, where a celebrity photo is juxtaposed with unrelated text. 
LLM-MRD resulted in a correct 'fake news' prediction. 
Conversely, the baseline MIMoE-FND\cite{liu2025modality} failed to detect this cross-modal inconsistency and misclassified the item as real.

\begin{figure}[htbp]
\centering
\includegraphics[width=0.8\linewidth]{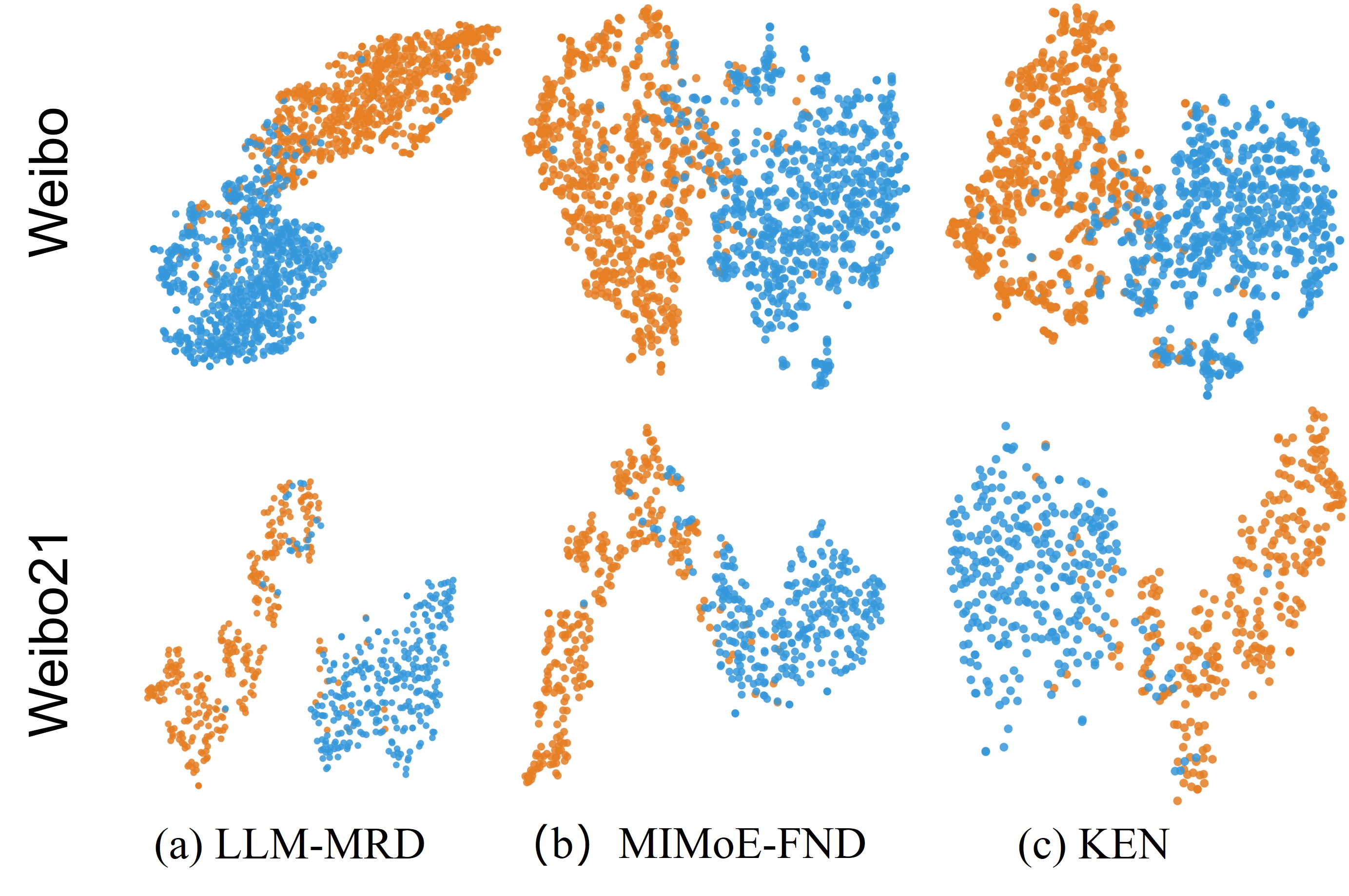}
\caption{T-SNE visualization of test set features. Same color dots indicate the same label.}
\label{fig:tsne_visualization}
\vspace{-6pt}
\end{figure}

\subsection{T-SNE Visualizations}
Figure \ref{fig:tsne_visualization} shows t-SNE visualizations of the features learned by LLM-MRD, MIMoE-FND\cite{liu2025modality}, and KEN\cite{zhu2025ken} on the Weibo and Weibo21 test sets.Compared to MIMoE-FND\cite{liu2025modality} and KEN\cite{zhu2025ken}, LLM-MRD produces fewer outliers in its fake news representations and exhibits less overlap between real and fake news embeddings.These findings further confirm the superior performance of LLM-MRD in multimodal fake news detection.

\indent
The feature distribution of LLM-MRD exhibits a completely different pattern, forming multiple clearly separated subclusters.This stands in stark contrast to the relatively single, concentrated clusters formed in the Weibo dataset.
This phenomenon suggests that the Weibo21 dataset contains multiple news events with widely varying topics.
The LLM-MRD model not only successfully distinguishes the authenticity of news, but also further captures deep semantic information at the event level, spatially distinguishing the features of different events.

\begin{table}[htbp]
    \centering
    \vspace{-12pt} 
    \caption{Comparative Reasoning efficiency analysis of LLM-MRD and baseline models}
    \label{tab:ablation_results}
    \resizebox{0.8\linewidth}{!}{%
    \begin{tabular}{lccc}
        \toprule
        \rowcolor{lightcyan}
        \textbf{Method} & \textbf{Params} & \textbf{Reason latency} & \textbf{Time Complexity} \\
        \midrule
        MIMoE-FND & 397M & 20.5ms & $O(k \cdot n_e \cdot (n^{2d}))$ \\
        KEN & 382M & 19.7ms & $O(L \cdot (n^{2d}))$ \\
        RaCMC & 424M & 18.8ms & $O(L \cdot (n^{2d}))$ \\
        GLPN-LLM & 7.07B & 158ms & $O(L_{\text{llm}} \cdot n_{\text{llm}}^{2d})$ \\
        \midrule
        \textbf{LLM-MRD} & \textbf{358M} & \textbf{17.2ms} & \textbf{$O(L \cdot (n^{d}))$} \\
        \bottomrule
    \end{tabular}
    }
    \vspace{-6pt} 
\end{table}

\subsection{Inference efficiency}
LLMs are key for fake news detection via strong reasoning, but direct deployment faces high compute and latency hurdles. Our LLM-MRD mitigates this with an efficient teacher–student framework. A powerful LLM \emph{teacher} performs deep reasoning offline, transferring it to a smaller \emph{student} via calibration distillation. The resulting student achieves efficient inference by operating without the LLM teacher during the deployment phase.

As shown in Table \ref{tab:ablation_results}, our student model (358M parameters), integrating three encoders, has a parameter size comparable to strong baselines like KEN\cite{zhu2025ken} (382M) and RaCMC\cite{yu2025racmc} (424M).

However, our framework's efficiency is twofold. First, it achieves faster reasoning latency (17.2ms)compared to these baselines (19.7ms and 18.8ms, respectively). Second, it avoids the prohibitive inference costs of methods relying on multi-billion parameter LLMs, such as GLPN-LLM\cite{hu2025synergizing} (7.07B parameters, 158ms latency).

This efficiency does not sacrifice accuracy. LLM-MRD surpasses all strong baselines, improving average accuracy by 5.19\%, F1-Fake by 6.33\%, and F1-Real by 5.63\% across three challenging datasets \cite{tong2025dapt}. These results show our framework effectively transfers a large teacher’s complex reasoning-derived knowledge to a deployable student architecture, yielding state-of-the-art performance while remaining efficient.

\section{Conclusion} In this paper, we introduce LLM-MRD, a novel teacher-student distillation framework designed to advance multimodal fake news detection. Our method is built upon three core components: Student Multi-view Reasoning, Teacher Multi-view Reasoning, and Calibration Distillation. This approach effectively aligns the student’s representations with the teacher's multifaceted reasoning across textual, visual, and cross-modal domains. A Multi-View Fusion module subsequently performs robust semantic integration for the final decision. Extensive experiments conducted on the Weibo, Weibo-21, and GossipCop benchmarks validate that LLM-MRD achieves consistent improvements in accuracy, robustness, and efficiency. This work offers a principled paradigm for transferring complex LLM reasoning-derived knowledge to compact architectures, enabling efficient and trustworthy detection systems.

\begin{credits}
\subsubsection{\ackname} This research was supported by the Finance science and technology project of Xinjiang Uyghur Autonomous Region(2023B01029-1, 2023B01029-2), the Outstanding Young Talent Foundation of the Xinjiang Uygur Autonomous Region of China (Grant No. 2023TSYCCX0043), and the National Natural Science Foundation of China (Grant No. 62266043).

\subsubsection{\discintname}
The authors have no competing interests to declare that are relevant to the content of this article.
\end{credits}

\end{document}